# *From Text to Transformation*: A Comprehensive Review of Large Language Models' Versatility

Pravneet Kaur*
Jamia Hamdard
New Delhi, India

Gautam Siddharth Kashyap* #
IIIT Delhi
New Delhi, India
officialgautamgsk.gsk@gmail.com

Ankit Kumar
IIT Bombay
Mumbai, India

Md Tabrez Nafis
Jamia Hamdard
New Delhi, India

Sandeep Kumar
MSIT
New Delhi, India

Vikrant Shokeen
MSIT
New Delhi, India

* Equal Contribution
# Corresponding Author

## ABSTRACT

This groundbreaking study explores the expanse of Large Language Models (LLMs), such as Generative Pre-Trained Transformer (GPT) and Bidirectional Encoder Representations from Transformers (BERT) across varied domains ranging from technology, finance, healthcare to education. Despite their established prowess in Natural Language Processing (NLP), these LLMs have not been systematically examined for their impact on domains such as fitness, and holistic well-being, urban planning, climate modelling as well as disaster management. This review paper, in addition to furnishing a comprehensive analysis of the vast expanse and extent of LLMs' utility in diverse domains, recognizes the research gaps and realms where the potential of LLMs is yet to be harnessed. This study uncovers innovative ways in which LLMs can leave a mark in the fields like fitness and wellbeing, urban planning, climate modelling and disaster response which could inspire future researches and applications in the said avenues.

## 1 INTRODUCTION

Large language models (LLMs) are basically artificial intelligence models architecture using deep learning and sophisticated algorithms that have revolutionized communication and comprehension by reshaping the way machines interpret and generate human-like language. They represent the progress in Natural Language Processing (NLP) by transforming interaction of people with technology, unravelling new possibilities across varied domains. Transformative power of deep learning harnessed via transformative architecture, which is trained on voluminous amounts of textual data and refined through human feedback, composes the core of LLMs. The practice of pre-training of LLMs combined with the integration of fine tuning and reinforcement learning from human feedback enables them to produce responses taking into account the context of input text, serving to ensure improvised and more efficient performance of LLMs [1]. Owing to this pre-training, applicability of LLMs extends beyond mere linguistic dexterity and has been explored in diverse domains spanning virtual assistance [2]–[4], code generation and comprehension [5]–[7], medical text analysis [8], [9], customer support automation [10]–[12], legal document analysis [13], [14], and education [15]–[17].

Additionally, accounting to the ability to absorb nuances of grammar, semantics and pragmatics, LLMs, which are essentially trained massive neural networks, are able to mimic human responses with astonishing accuracy in a diverse array of tasks. LLMs have emerged as transformative tools with user friendly design that encourages their operation of people with non-technical backgrounds as well. The expanse of its utility can be depicted by its widespread deployment in varied arenas as discussed further. In the sphere of NLP, models like Bidirectional Encoder Representations from Transformers (BERT) have exhibited exemplary performance in tasks encompassing language comprehension and sentiment analysis [18]. These models assist in effectively capturing contextual information paving way to state-of-the-art results in NLP benchmarks. Further, another domain significantly impacted by LLMs is Information Retrieval (IR), where designing of models like T5 (Text-To-Text Transfer Transformer) depicted its utility in unification of several NLP tasks incorporating question answering thereby improving IR systems.

Furthermore, a yet another beneficiary of LLMs is the field of machine translation where LLMs have played an indispensable role in breaking down language barriers. Their effectiveness can be showcased by their significant contributions in facilitating multilingual communication entailing the exploration of cross-lingual applications via language-agnostic BERT sentence embeddings [19]. Even the healthcare and biomedical research and industry have not remained untouched by the transformative influence of LLMs which can be well exemplified by models like BioBERT [20] which has been particularly pre-trained on biomedical text mining. It assists in tasks such as entity recognition and relation extraction thereby contributing to enhancement in biomedical research [20]. The efficacy of LLMs in extracting medical information from electronic health records consequently streamlining healthcare data analysis can be further highlighted by a study conducted by [21].

Various studies evince the successful deployment of LLMs tailored to be deployed in the legal field as well where LLMs conferred benefits ranging from document summarization, contract analysis to legal research. LegalBERT introduced by [22], demonstrated the adaptability of language models in legal text comprehension which eased the otherwise tedious task of legal document understanding and contract analysis. Additionally, LLMs also simplified the summarization of legal texts and extraction of structured information from legal documents thereby contributing to more efficient legal processes by optimizing the process of contract review and analysis [23], [24]. Education and intelligent tutoring systems comprise of yet another sector where LLMs have left a mark. They have not only facilitated the development of intelligent tutoring systems, educational chatbots, and automated grading systems but they have upgraded the learning experience to an extraordinary level. This can be well illustrated by GPTune [25] that optimizes LLMs for educational applications by focusing on hyperparameter tuning for transformer models [25]. Further, a research by [26] delved into how the employment of LLMs in providing personalized feedback to students to promote individualized instruction and enrich learning experience. Furthermore, LLMs are being explored by researches as well to investigate their proficiency in accurate processing of vast amounts of literature, ability to generate sensible summaries and even hypothesis proposal. A study by [27] introduced SciBERT to examine the use of LLMs for production of scientific explanations, showcasing its potential in knowledge dissemination and hypothesis generation. The potential of LLMs has also been harnessed in the realm of computer vision and image understanding which is emphasized by recent work on ImageBERT [28] that showcased LLMs' extension to tasks beyond text, highlighting their capability of learning joint representations of images and text for image classification and retrieval.

An arena where LLMs have showcased remarkable performance is code generation and programming assistance which can be clearly evinced by CODEBERT, a model pre-trained on extensive code-text corpus, demonstrating the utility of LLMs in code-associated tasks [29]. Under the umbrella of beneficiaries of LLMs falls the field of finance and sentiment analysis as well where analysis of financial data and sentiment in market-related information, financial news and reports smoothened the process of several finance-associated tasks like prediction of stock market volatility and gain of valuable insights for investment decisions and risk management [30], [31]. RoBERTa [32], a robustly optimized BERT approach represents the advancements in NLP tasks including sentiment analysis in social media data, represents the LLMs-assisted leaps taken in the arena of social media analysis. Furthermore, the advent of language models like GPT-3.5 has proven to be groundbreaking. Its widespread application in multitudinous domains, where it serves to provide valuable insights, aid in automation and generates innovative solutions, is a testament to its versatility and adaptability. Chatbots powered by such models have become integral in providing instant and contextually relevant responses. These chatbots are being extensively employed in fields such as content creation and journalism to generate news articles, blogs and other written content at an accelerated pace. The cited studies clearly underscore the depth and width of recent advancements made possible via the transformative influence of LLMs and how they have become indispensable across multiple sectors, perpetually shaping and upgrading applications in NLP, IR, healthcare, legal analysis, education, scientific research, computer vision, social media analysis, code generation, finance, customer service and content creation.

One of the primary contributions of the study is to provide a thorough examination of the potential applications of LLMs in relatively unexplored arenas, namely climate modelling, urban planning, disaster response and fitness and meditation aided holistic well-being. The absence of a comprehensive review addressing intersection of LLMs with these arenas motivated our in-depth analysis of the prevailing expanse and untapped potential of LLMs in this study. The integration of LLMs into these realms presents an unprecedented opportunity to revolutionize the existence of humans as a society and not just on an individual level by facilitating urban renewal, enhancing disaster management and response and incorporating climate modellling to provide a better living standard with well-preparedness for any disastrous calamity in addition to aiding in holistic well-beings of individuals. Moreover, the environment also stands to benefit from the deployment of LLMs in climate modelling and disaster management by enabling more informed decision-making, facilitating conservation efforts, and promoting sustainable development practices. The paper serves as a foundational resource for scholars, practitioners, and enthusiasts interested in understanding the symbiotic relationship between LLMs and personal well-being practices, LLMs and climate modelling, LLMs and urban planning as well as LLMs and disaster response. Via this study we aim to delve into the following research questions and attempt to answer them in further sections in order to pave way for future researches in the same.

- **RQ1:** How can LLMs be leveraged to optimize both fitness and meditation practices, ensuring personalized guidance and support tailored to individual needs and preferences, thereby promoting holistic well-being?
- **RQ2:** In what ways can LLMs analyze urban data and propose innovative solutions for sustainable urban development, and how do these approaches compare to

- traditional urban planning methods in terms of efficiency and effectiveness?
- **RQ3:** How can LLMs leverage historical climate data, atmospheric conditions, and greenhouse gas emissions to develop advanced climate models, and what are the potential implications for improving the accuracy of climate predictions and informing climate change mitigation and adaptation strategies?
- **RQ4:** How can LLMs integrate real-time data sources and predictive modeling techniques to improve the accuracy and timeliness of disaster risk assessments, enabling more effective emergency response planning and resource allocation?

The paper is as follows in the next section we will see the working of LLMs. In Section 3, we will discuss the LLMs lifecycle. In Section 4, the methodology for literature review is presented. In Section 5, the studies that are relevant to the scope and research questions of the paper are cited. In Section 6, the LLMs toolstack is presented. In Section 7, the comparisons of LLMs toolstack is presented. In Section 8, the thoughts of the paper is presented and we conclude the paper in Section 9 with some conclusions.

## 2 WORKING OF LLMS

The deployment of LLMs has revolutionized various industries. The effectiveness of these models lies in their ability to capture and generate human-like language through probabilistic predictions. One remarkable feature of LLMs is their capacity for transfer learning. Despite their successes, ethical concerns surround the use of LLMs. LLMs, such as GPT-3.5, operate on the principles of deep learning, specifically within the domain of NLP. The ability to transfer knowledge acquired from one domain to another reduces the need for extensive task-specific training, making LLMs efficient and cost-effective for a wide array of applications. Their prowess in language translation has led to improved cross-language communication, breaking down language barriers. Moreover, LLMs have been employed in educational settings for personalized tutoring, adapting to individual learning styles and providing tailored feedback. During training, the model learns to capture the intricate patterns, contextual nuances, and syntactic structures inherent in language. These models are built upon neural network architectures, employing transformers that facilitate the processing of sequential data, like language. The architecture comprises layers of attention mechanisms that enable the model to assign varying degrees of importance to different parts of the input sequence, facilitating the understanding of long-range dependencies in language. The models excel in content creation, generating human-like articles, stories, and poetry (Fig. 1). This probabilistic approach results in diverse and contextually appropriate responses, simulating the variability and creativity found in human language. In natural language understanding tasks, they exhibit a nuanced comprehension of user queries, enabling more accurate and context-aware responses in virtual assistants, customer support chatbots, and search engines. The transformer architecture, a key component, allows for the contextual understanding of words within a given context, enabling the model to generate coherent and contextually relevant responses. The training process involves exposing the model to vast datasets, comprised of diverse textual information from the internet, books, and other sources. The potential for biased outputs, the reinforcement of existing stereotypes presents in the training data, and the generation of inappropriate or harmful content raise ethical considerations. Attention mechanisms also play a crucial role in handling contextual information, as the model considers the entire context when generating each word, rather than relying solely on preceding words. Pre-trained on massive datasets, these models can be fine-tuned for specific tasks with comparatively smaller datasets. The term "large" denotes the extensive scale of parameters, often numbering in the billions, which contribute to the model's capacity to comprehend and generate human-like text. The training process involves adjusting the millions or billions of parameters through backpropagation and gradient descent, minimizing the difference between the model's predictions and the actual target sequences. This characteristic contributes to the model's ability to grasp the broader context of a conversation or text, thereby enhancing the quality of generated content. However, it also introduces challenges, such as the potential generation of biased or inappropriate content, requiring careful consideration and moderation in real-world applications. During inference, the model produces a distribution of probable next words, with the most likely word chosen based on sampling techniques or other strategies.

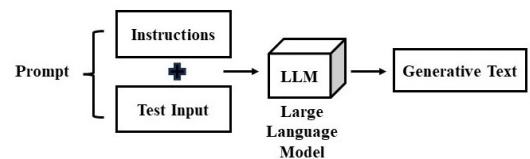

**Fig. 1.** Working of LLMs

## 3 LLMS LIFECYCLE

The LLMs lifecycle is a multifaceted and meticulous process essential for the development, deployment, and ethical use of sophisticated language models. It begins with a clear definition of the problem at hand and the establishment of realistic goals, setting the foundation for subsequent stages (refer to Fig. 2). Understanding specific requirements and identifying the target audience are crucial components of this initial phase, providing guidance for the development team as they work towards aligning the model with desired outcomes. Moving beyond problem definition, the lifecycle advances to the pivotal stages of data collection and preprocessing. The quality of data proves imperative for the success of any language model, necessitating the assembly of diverse and representative datasets that encapsulate the variability present in real-world scenarios. The raw data, often riddled with noise, undergoes preprocessing involving cleaning, tokenization, and normalization to ensure it is in a suitable format for effective training and evaluation. This meticulous data preparation becomes fundamental to the model's ability to generalize well to new, unseen examples. Critical decisions in the lifecycle arise during the stages of model selection and architecture design. Depending on the nature of the task at hand, developers must carefully choose appropriate model architecture, ranging from traditional Recurrent Neural Networks (RNNs) and Long Short-Term Memory Networks (LSTMs) to more recent innovations like transformer architectures. The selected model must possess the requisite complexity and capacity to capture nuanced patterns within the data. This phase requires a delicate balance, aiming for model expressiveness without succumbing to overfitting, ensuring the model's adaptability to a diverse range of language patterns. With the architecture in place, the model transitions to the training phase, a critical step where it learns from the prepared data by adjusting its internal parameters to minimize the difference between predicted and actual outputs. Training involves optimization algorithms such as stochastic gradient descent, and the duration varies depending on the size and complexity of the model. Rigorous validation during training is indispensable to prevent overfitting to the training data and offers insights into the model's generalization capabilities. Following successful training, hyperparameter tuning becomes the focus—a process of fine-tuning parameters like learning rate, batch size, and regularization strength. This iterative phase aims to strike a balance, enhancing the model's performance without compromising its ability to generalize effectively to new, unseen data. The success of hyperparameter tuning significantly influences the overall performance of the language model.

Post-training, the model undergoes evaluation and validation using separate datasets. Metrics such as accuracy, precision, recall, and F1 score are calculated to assess its performance. If the model falls short of predefined benchmarks, adjustments may be made, and additional training cycles might be necessary. This iterative approach ensures that the model meets specified performance criteria, instilling confidence in its reliability and effectiveness. Once successfully trained and validated, the model enters the deployment phase, where it is integrated into the target system or application. This phase demands meticulous attention to considerations such as scalability, latency, and resource utilization to ensure the model can handle real-time requests efficiently. Deployment marks the transition from development to real-world application, and thorough testing is paramount to ensure the model performs as expected in diverse and dynamic environments. The post-deployment phase introduces essential elements of monitoring and maintenance. Continuous monitoring is critical to detect any performance degradation or changes in the data distribution that may impact the model's accuracy. This phase also includes regular updates and maintenance to address challenges such as concept drift, where the model's assumptions become outdated over time. Ethical considerations and bias mitigation remain integral throughout the entire lifecycle. Scrutinizing the training data for biases and conducting regular audits help ensure fair and responsible predictions, particularly regarding certain demographic groups. This commitment to ethical practices extends through deployment, monitoring, and maintenance, promoting fairness, transparency, and accountability in the model's usage. User feedback plays a pivotal role in refining and improving the model post-deployment. Continuous iteration based on user experience and feedback contributes to the model's ongoing relevance and effectiveness in addressing real-world challenges, enhancing its capabilities to meet evolving user needs. A robust feedback loop ensures that the model remains a valuable tool in various applications.

Documentation stands as a fundamental aspect of the LLMs lifecycle, capturing crucial details about problem definition, dataset, model architecture, training process, and deployment considerations. Comprehensive documentation facilitates knowledge transfer within the development team and supports future maintenance and updates, ensuring that the model remains well-understood and effectively managed. As technology evolves, models may become outdated or face challenges in keeping up with emerging trends. The retirement or replacement of models becomes necessary to meet new requirements and address evolving challenges effectively. This less frequent but crucial phase represents a natural progression in the lifecycle as developers strive to ensure that their models remain state-of-the-art and relevant to the current linguistic landscape. The LLMs lifecycle encapsulates a dynamic and iterative process, emphasizing responsible development, ethical use, and continuous improvement to meet the ever-changing demands of language processing in the digital era.

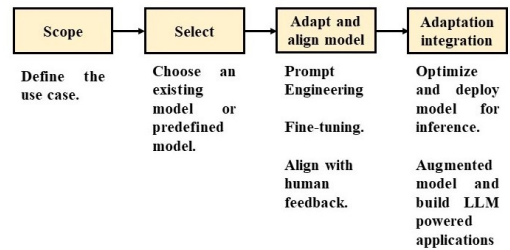

**Fig. 2.** LLMs Lifecycle

## 4 METHODOLOGY FOR LITERATURE REVIEW

In the context of the previously described research questions, the technique for carrying out a literature review entails a methodical approach to locating, compiling, and evaluating pertinent scholarly material on the application of LLMs across diverse domains. Finding important themes and search phrases about LLMs and their uses in fields like urban planning, climate modeling, fitness, and disaster relief is the first step in the process. "Large Language Models", "GPT", "BERT", "fitness", "well-being", "urban planning", "climate modeling", "disaster response", and associated topics are examples of possible search queries. A thorough search strategy is developed to find pertinent literature by utilizing academic databases, scholarly journals, conference proceedings, and reliable online sources. To ensure inclusivity and refine the search results, complex search algorithms and boolean operators are utilized. Sources that have been identified are carefully assessed for quality, relevance, and trustworthiness based on a variety of factors, including publication date, author expertise, study technique, and alignment with the abstract's stated research aims. To extract important insights and conclusions, a selection of literature is then arranged and summarized. In order to determine areas of agreement, research gaps, and future research directions, the synthesized literature is examined. Innovative uses of LLMs in fields like fitness, well-being, urban planning, climate modeling, and disaster response are given particular consideration. The methodology ends with a critical analysis of the literature review procedure and its implications for future studies, along with suggestions for additional study and useful uses of LLMs in the domains that have been discovered. The study intends to provide a thorough examination of LLM adoption across a variety of sectors while highlighting areas for further research and innovation through the use of a rigorous and methodical approach.

## 5 LITERATURE REVIEW

In this section, we provide a brief overview of the extent of exploration done till date in general and in related fields as our study.

### 5.1 General Overview

The utilization of LLMs has witnessed an astounding surge across diverse fields. LLMs have demonstrated remarkable capabilities in the comprehension and processing of natural language, as seen in research exploring sentiment analysis, entity recognition and language translation [18], [33], [34]. Researchers also dug into the potential of LLMs in the field of content generation to establish that LLMs can make impactful contribution in text creation, article summarization and even code generation. In addition, LLMs effectively serve as a foundation for conversational agents and chatbots as well thereby facilitating enhanced human-computer interactions in customer support and virtual assistance [35], [36]. Legal domain has also witnessed the integration of LLM for legal document analysis, contract review and summarization of legal texts [37]. Additionally, LLMs have become pivotal in education technology as well, assisting in tasks like automated grading and content generation for educational purposes [38], [39]. The influence of LLMs further extends to the financial domain as well where it facilitates financial data and sentiment analysis [31]. LLMs have indeed proven to be not just valuable tools but rather dynamic collaborators that have been influencing and reshaping a myriad of fields with their language – driven capabilities as discussed above.

### 5.2 Fitness and Well-Being

In the realm of wellness, LLMs have majorly been explored in terms of healthcare. This exploration encompasses medical text analysis, aiding in tasks such as information extraction and summarization of medical literature [21], [40]. Further, LLMs are also involved in patient-care communication via analysis of doctor-patient interaction, generation of health-related information and patient-friendly content [41]. Furthermore, they serve to provide mental health support via providing therapy, suggestions for mental well-being and sentiment analysis of mental health related-texts [42], [43]. Moreover, LLMs contribute towards creation of conversational agents for healthcare that offer health-related queries, medication reminders and provide general wellness guidance [44], [45]. A lot more is yet to be explored in this domain of fitness and well-being. LLMs can be leveraged for revolutionizing fitness and meditation practices by utilizing individual feedback and emotional responses. By analyzing the personal health data LLMs can tailor personalized workout plans and nutrition strategies based on each individual's unique needs and preferences. This personalized approach cannot only cater to fitness outcomes but shall also foster sustained adherence and engagement ultimately leading to improved physical health and overall well-being. A recent study conducted by [11] depicted the effectiveness and acceptability of coaching messages generated by LLM, particularly ChatGPT, in comparison to those produced by humans for behavioral weight loss purposes. The findings suggest that LLM-based chatbots have the potential to offer personalized and innovative messages, albeit sometimes perceived as slightly formulaic and less authentic compared to human-generated ones. These pitfalls can be overcome by integrating more sophisticated natural language understanding algorithms and incorporating additional user data, such as historical feedback and individual preferences which could enhance the response by generating more nuanced, contextually appropriate, and personalized coaching messages. Furthermore, [46] introduced PlanFitting, a conversational system aimed at assisting users in developing personalized exercise plans. Although the primary focus was on exercise planning, the study highlights the potential of LLMs in furnishing personalized, actionable, and evidence-based plans custom-made for individual circumstances and preferences. In a complementary perspective, [47] critically evaluated the efficacy of exercise prescriptions generated by OpenAI's GPT-4 model. The study evaluated the model's ability to construct personalized exercise programs for individuals with varied health conditions and fitness goals on the basis of singular, initial interaction. Although the findings indicated that models, inclusive of GPT-4, can produce general safety-conscious exercise programs, it suggests they may lack precision in catering to individual health condition and goals. The study highlights the role of LLMs technologies as supplemental tools in exercise prescription, particularly for enhancing accessibility, but warns against substituting them for personalized and condition-specific prescriptions provided by healthcare and fitness professionals. It emphasized the need for a deeper exploration of more interaction use of LLMs models and real-time physiological feedback. By amalgamating insights from these studies, we can envision a similar approach applied to meditation practices. LLMs have the capacity to interpret user feedback and emotional cues during guided meditation sessions, modifying the meditation content in real-time to address specific emotional states and promote emotional well-being, stress reduction, and mindfulness. While challenges such as authenticity and formulaic responses need to be overcome, the demonstrated preliminary feasibility in weight loss coaching suggests the potential for LLMs to enhance meditation practices by providing personalized guidance and support tailored to individual needs and preferences.

### 5.3 Urban Planning

Attributing to the capability to analyze vast datasets and generate innovative solutions, LLMs present remarkable opportunities for urban planning and can be deployed to analyze urban data and propose sustainable development solutions. By processing diverse urban datasets encompassing demographics, transportation, land use and the environment, LLMs can generate insights into urban dynamics, population trends in addition to infrastructure needs [48], [49]. [48] demonstrated a novel approach of fine-tuning LLMs specifically for urban renewal tasks resulting in enhanced efficiency and accuracy of IR and generation in this domain. Their research introduced a method that considerably improvised LLM's performance on urban renewal knowledge QA tasks, showcasing the potential for optimizing LLMs in urban planning contexts. In alignment with this, the results of another study by [49] depicted the superior performance of UrbanCLIP in urban region profiling, showcasing its potential to enhance urban planning processes through the integration of textual modality with image data. Their study evaluated the performance of UrbanCLIP in predicting urban indicators across major Chinese metropolises, highlighting its effectiveness in generating detailed textual descriptions for satellite images and its potential for advancing urban planning practices. Traditional urban planning methods come across as relatively less efficient and more time-consuming as they rely on static datasets and manual analysis thereby leading to limited understanding and delayed decision-making processes. While LLM-driven approaches can offer real-time analysis of dynamic urban data streams thereby enabling proactive decision-making and fostering creativity and innovation in addressing complex urban challenges [50]. Insights provided by the study by [50] can enhance our understanding of the adaptability and effectiveness of LLMs in participatory urban planning. Their work emphasized the ability of LLMs to outperform human experts and state-of-the-art methods in generating inclusive and adjustable solutions through role-play, collaborative generation, and feedback iteration. By focusing on community-level land-use tasks, their study demonstrated the potential for LLMs to facilitate collaborative decision-making processes in urban planning contexts. Moreover, by leveraging recent advancements like Urban Generative Intelligence (UGI) [51], LLMs can simulate systems and interact with embodied agents to manage urban complexity effectively. UGI was proposed by [51] as a transformative platform that integrates LLMs into urban systems, aiming to address the complex challenges of urban environments through multidisciplinary approaches. Their study introduced a foundational platform for leveraging LLMs in addressing urban issues such as traffic congestion, pollution, and social inequality, highlighting the potential for simulating complex urban systems and managing urban complexity in a multidisciplinary manner. Despite the persisting challenges, incorporation of numerical data and interactions with simulations, solutions like TrafficGPT [52] demonstrate the potential of combining LLMs with domain-specific models to provide decision support for urban transportation management. TrafficGPT integrates LLMs with traffic foundation models to advance traffic management practices. Their research introduced a fusion approach that empowers LLMs to view, analyze, and process traffic data, providing decision support and interactive feedback for urban transportation system management. By seamlessly intertwining LLMs with traffic expertise, TrafficGPT showcases the potential for LLMs to enhance stakeholder engagement and improve the efficiency of urban infrastructure in real-world traffic scenarios. In conclusion, LLM-driven urban planning approaches offer efficiency, effectiveness and innovation in comparison to traditional methods thereby paving way for sustainable and resilient cities of the future.

### 5.4 Climate Modelling

LLMs are emerging as valuable tools in the realm of climate modelling as well. Employing their capacity to process and contextualize extensive datasets encompassing historical climate records, atmospheric dynamics, and greenhouse gas emissions. LLMs can detect complex relationships and trends that can assist in the development of advanced climate models by integrating climate data from diverse sources. Recent research efforts, such as that by [53], have focused on evaluation of the effectiveness of LLMs in communicating climate change information. This study underscored the potential for LLMs to bridge the gap between complex scientific data and public understanding, empowering individuals and communities to engage in climate change mitigation and adaptation efforts. Furthermore, studies like the one conducted by [54] highlight the role of LLMs, such as ChatGPT, in synthesizing data and mobilizing public awareness about climate change. These models streamline the communication of climate-related information thereby expediting informed decision-making and action. Furthermore, a study by [55] proposed novel approaches to grant access to recent and precise climate-related data which can enhance the accuracy of climate predictions. This advancement not only caters to a critical limitation in traditional modeling approaches but also contributes to the refinement of predictive models. In addition, tailored-LLMs such as Arabic Mini-ClimateGPT introduced by [56], demonstrate the potential for language adaptation in engaging specific communities on climate change issues. By overcoming language barriers, these models spread awareness and encourage participation in sustainability efforts. Further, to address challenges pertaining to climate change question-answering tasks, [57], enhanced LLMs with access to reliable IPCC reports. This augmentation further improvises the accuracy and reliability of responses, promoting informed decision-making and public discourse on climate-related issues. Additionally, [58] utilized tools like ClimSight to explore the democratization of climate information by leveraging LLM capabilities to provide localized climate services worldwide. By integrating geographical information along with model simulations, these tools offer actionable insights to individuals, empowering proactive responses to climate change at local levels. In essence, LLMs offer a multifaceted approach to climate modeling, incorporating data analysis, predictive modeling, communication, language adaptation, and IR. These capabilities position LLMs as valuable assets in furthering climate science and fostering informed decision-making for climate change mitigation and adaptation on a global scale.

## 5.5 Disaster Response

Recent studies have demonstrated how LLMs can be deployed for upgrading disaster response efforts by integrating real-time data sources and predictive modeling techniques. [59] leveraged LLMs for fabrication of DisasterResponseGPT based on an algorithm designed to swiftly generate valid plans of action in disaster scenarios. This approach facilitates plan development and allows for real-time adjustments during execution thus enhancing the agility and effectiveness of disaster response operations. Further, [60] explored the role of LLMs in radiation emergency response, highlighting the importance of expert guidance and domain-specific training to integrate LLMs into decision support frameworks effectively. Furthermore, another study by [61] introduced FloodBrain, a tool that utilizes LLMs to generate detailed flood disaster impact reports, facilitating coordination of humanitarian efforts in the aftermath of floods. Moreover, [62] assessed the geographic and geospatial capabilities of Multimodal Large Language Models (MLLMs), providing insights into their strengths and challenges in visual tasks within geographic domains. Additionally, [63] aimed to converge Geospatial Artificial Intelligence (GeoAI) and disaster response, showcasing GeoAI's applications via social media data analysis during Hurricane Harvey. Their study aises in identification of opportunities and challenges for leveraging GeoAI in disaster response, highlighting its potential to improve situational awareness and decision-making. Additionally, [64] proposed combining digital twins and chatbots using advanced language models to enhance smart cities' resilience to disasters, in order to provide decision support interfaces to disaster managers and augment collaborative response efforts. Overall, these studies demonstrate the versatility and effectiveness of LLMs in disaster response by facilitating rapid plan generation, real-time data analysis, and decision support. By harnessing the power of LLMs, disaster management agencies can improve their capabilities to prepare for, respond to, and recover from disasters, ultimately saving lives and reducing the socio-economic impact on affected communities.

## 6 LLMS TOOLS STACK

The comprehensive toolkit for LLMs covers web scraping tools, deep learning frameworks, pre-trained models, deployment solutions, and ethical considerations. Components like Beautiful Soup[1], TensorFlow[2], BERT, and Docker play[3] crucial roles in various stages, ensuring efficiency and ethical use. Continuous integration is managed by Jenkins[4] and Travis CI[5], providing a holistic approach to the development and deployment of sophisticated language models.

### 6.1 Data Collection and Preprocessing Tools

In the domain of LLMs data collection and preprocessing, pivotal tools ensure the quality and suitability of input data (Fig. 3). Python libraries like Pandas[6] and NumPy[7] serve as the foundation for data manipulation, offering versatile capabilities in handling structured data and performing essential numerical operations. The Natural Language Toolkit (NLTK) plays a crucial role in text processing and analysis, providing functionalities such as tokenization, stemming, and part-of-speech tagging. These tools collectively empower developers to prepare datasets for training, ensuring data is cleansed, tokenized, and normalized for subsequent model lifecycle stages. By navigating the intricacies of diverse datasets, practitioners extract valuable insights and patterns, contributing to the robustness and effectiveness of LLMs in various NLP tasks. The integration of Pandas, NumPy, and NLTK facilitates a comprehensive approach, enhancing the overall data preparation process for the development and deployment of sophisticated language models.

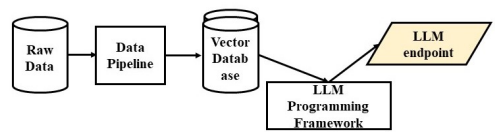

**Fig. 3.** Data Collection

### 6.2 Model Development and Training Tools

In the sphere of LLMs development and training, a versatile toolkit empowers developers to create sophisticated neural networks and optimize their performance. TensorFlow and PyTorch[8], prominent deep learning frameworks, offer high-level abstractions for crafting intricate neural architectures and efficient training on diverse datasets. The user-friendly Keras API complements these frameworks, simplifying the model development process. Hugging Face Transformers[9] is instrumental, enabling exploration and fine-tuning of pre-trained transformer models like BERT and GPT. Noteworthy too is AllenNLP[10], built on PyTorch, which specializes in NLP tasks, providing pre-built models and tools for custom model construction. This comprehensive set of tools equips developers to navigate the intricacies of model architecture design and training, facilitating the creation of powerful LLMs adept at addressing a wide range of natural language understanding and generation challenges. Fig. 4 shows the block diagram related to model development and training tools.

---

[1] https://beautiful-soup-4.readthedocs.io/en/latest/
[2] https://www.tensorflow.org/
[3] https://labs.play-with-docker.com/
[4] https://www.jenkins.io/
[5] https://www.travis-ci.com/
[6] https://pandas.pydata.org/
[7] https://numpy.org/
[8] https://pytorch.org/

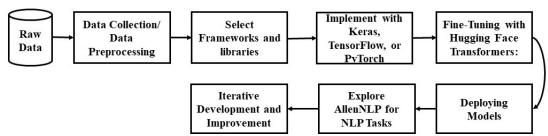

**Fig. 4.** Model Development and Training Tools

### 6.3 Hyperparameter Tuning Tools

In the quest to optimize LLMs, hyperparameter tuning tools play a crucial role in augmenting performance and generalization. Optuna[11], an influential open-source framework, excels in automating hyperparameter optimization with efficient search algorithms, enabling systematic exploration of hyperparameter spaces. Another noteworthy tool, Hyperopt[12], employs Bayesian optimization algorithms to navigate these spaces effectively, aiding in the identification of optimal configurations. These tools simplify the iterative fine-tuning of parameters like learning rates and batch sizes, striking a delicate balance between overfitting and underfitting. Automation in the search for optimal hyperparameters enhances model accuracy and responsiveness. This process is integral to refining LLMs, ensuring robust performance across diverse datasets and real-world scenarios, ultimately contributing to their effectiveness in a spectrum of NLP tasks. Fig. 5 shows the block diagram related to hyperparameter tuning tools.

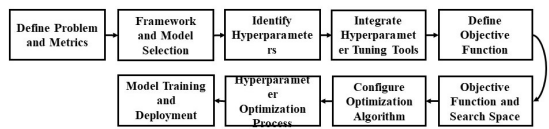

**Fig. 5.** Hyperparameter Tuning Tools

### 6.4 Evaluation and Validation Tools

In the expansive domain of LLMs, the indispensability of evaluation and validation tools is underscored, crucial for gauging model performance and ensuring effectiveness across diverse tasks. Scikit-Learn emerges as a cornerstone in this realm, offering a versatile machine learning library equipped with a comprehensive array of metrics and tools for model evaluation. From fundamental metrics like accuracy to nuanced measures such as precision, recall, and F1 score, Scikit-Learn facilitates a thorough assessment of model effectiveness. Expanding the toolkit, NLTK, renowned for its text processing capabilities, extends its utility to model evaluation. It provides tools like BLEU[13] score calculations, particularly valuable in assessing performance, especially in machine translation tasks. These integrated tools empower developers to rigorously validate LLMs, employing a variety of metrics for a holistic evaluation of their performance in specific NLP tasks. Beyond meeting predefined benchmarks, the integration of these evaluation and validation tools offers valuable insights for fine-tuning and iteration, contributing to the continuous improvement of LLMs. This iterative refinement enhances their ability to comprehend and generate human-like text across a spectrum of applications, solidifying their efficacy in the evolving landscape of NLP. Fig. 6 shows the block diagram related to evaluation and validation tools.

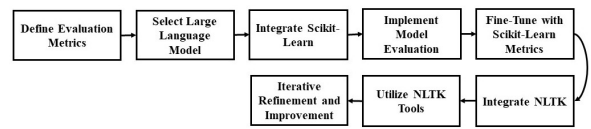

**Fig. 6.** Evaluation and Validation Tools

### 6.5 Deployment Tools

Deployment tools are pivotal in bridging the gap from LLMs development to real-world applications, ensuring seamless integration into production environments. TensorFlow Serving[14] and TorchServe play[15] crucial roles, providing dedicated APIs for model serving, versioning, and scalability, streamlining the deployment pipeline from training to practical implementation. Open Neural Network Exchange (ONNX[16]) offers an open standard for machine learning model representation, promoting interoperability between frameworks and simplifying deployment across diverse environments. Addressing challenges related to latency and resource efficiency, these tools ensure language models can promptly and reliably handle requests in production scenarios. Their incorporation not only simplifies deployment processes but also contributes to the robust and efficient utilization of LLMs, enhancing their accessibility and effectiveness across a wide range of applications, including chatbots and natural language interfaces, in various industries. Fig. 7 shows the block diagram related to deployment tools.

---

[9] https://huggingface.co/docs/transformers/en/index
[10] https://allenai.org/allennlp
[11] https://optuna.org/
[12] http://hyperopt.github.io/hyperopt/
[13] https://www.thepythoncode.com/article/bleu-score-in-python
[14] https://www.tensorflow.org/tfx/guide/serving
[15] https://pytorch.org/serve/
[16] https://onnx.ai/

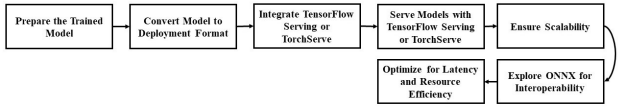

**Fig. 7.** Deployment Tools

### 6.6 Monitoring and Maintenance Tools

Monitoring and maintenance tools are crucial elements throughout the lifecycle of LLMs (refer to Fig. 8), ensuring continuous performance, reliability, and adaptability in real-world scenarios. TensorBoard[17], an integral part of TensorFlow, supplies essential visualization tools for monitoring and troubleshooting models during their training phase, providing valuable insights into metrics, model architectures, and potential issues. In production settings, the combined use of Prometheus[18] and Grafana[19] offers robust monitoring and alerting capabilities, diligently tracking system and model performance metrics to proactively detect and address anomalies. These tools empower practitioners to observe the well-being of deployed models, guaranteeing optimal responsiveness. Consistent monitoring is indispensable for recognizing performance degradation or alterations in data distribution, allowing for timely interventions and updates. This stage is particularly significant for tackling challenges like concept drift, where the model's assumptions become outdated over time. Collectively, these monitoring and maintenance tools contribute to the sustained effectiveness of LLMs, facilitating their adjustment to evolving conditions, user requirements, and ethical considerations across their operational lifecycle.

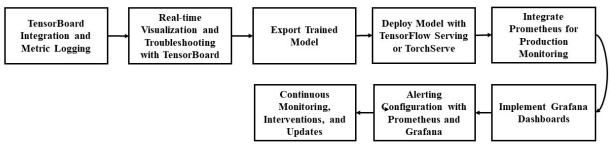

**Fig. 8.** Monitoring and Maintenance Tool

### 6.7 Ethical Considerations and Bias Mitigation Tools

Considering ethical implications and mitigating biases are critical components in the development of LLMs, and various tools play a crucial role in fostering responsible AI practices. The AI Fairness 360 (AIF360[20]) toolkit stands out as an open-source solution, providing metrics and algorithms for detecting and rectifying bias in machine learning models. It offers a comprehensive set of tools to assess and mitigate bias across various demographic groups, ensuring fairness and equity in the model's predictions. Another valuable tool, Fairness Indicators[21], is a TensorFlow solution designed to evaluate and improve model fairness by offering visualizations and metrics. Together, these tools empower developers to conduct thorough ethical audits, promoting the responsible and unbiased utilization of LLMs. Regular scrutiny and efforts to mitigate biases contribute to models not only performing effectively but also aligning with ethical standards, fostering transparency, fairness, and accountability in their deployment across a variety of applications and user demographics. Fig. 9 shows the block diagram related to ethical implications and mitigating biases.

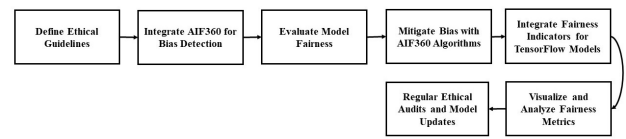

**Fig. 9.** Ethical Considerations and Bias Mitigation Tools

### 6.8 User Feedback and Iteration Tools

In the LLMs lifecycle, user feedback and iteration tools play a crucial role, allowing developers to enhance models based on real-world user experiences (refer to Fig. 10). Incorporating user analytics platforms like Google Analytics[22] or custom in-app analytics facilitates the gathering of valuable insights and usage patterns. This feedback loop directly understands user satisfaction, identifies challenges, and guides iterative improvements. Application-specific feedback mechanisms, such as in-app feedback forms or sentiment analysis on user reviews, provide granular insights into user sentiment and preferences. These tools enable developers to iteratively refine models, address usability issues, improve natural language understanding, and tailoring models to evolving user needs. The ongoing integration of user feedback into the development process ensures that LLMs remain responsive to real-world challenges, delivering a user experience aligned with expectations and contributing to the continuous evolution and optimization of the models for diverse applications and user scenarios.

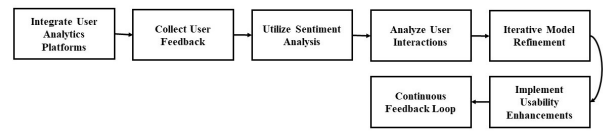

**Fig. 10.** User Feedback and Iteration Tools

### 6.9 Documentation and Knowledge Transfer Tools

---

[17] https://www.tensorflow.org/tensorboard
[18] https://prometheus.io/
[19] https://grafana.com/
[20] https://aif360.res.ibm.com/

[21] https://www.tensorflow.org/tfx/guide/fairness_indicators
[22] https://marketingplatform.google.com/about/analytics/

Documentation and tools for knowledge transfer are fundamental to ensuring the transparency, collaboration, and maintainability of LLMs throughout their lifecycle (refer to Fig. 11). Jupyter Notebooks[23] prove invaluable in this regard, offering a collaborative platform for creating and sharing documents that seamlessly blend live code, visualizations, and narrative text. These notebooks serve as a comprehensive record of the development process, aiding in the comprehension of model architectures, training procedures, and evaluation metrics. Similarly, Swagger[24]/OpenAPI[25] tools play a crucial role in standardizing and documenting APIs, promoting clear communication among different components of the model stack. Thorough documentation is essential for knowledge transfer within development teams, enabling new team members to grasp the intricacies of the model, its objectives, and operational requirements. These tools collectively contribute to the dissemination of knowledge, fostering collaboration, and supporting future maintenance and updates by providing a detailed record of the decisions, processes, and considerations that influence the development and deployment of LLMs.

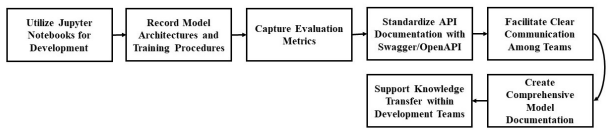

**Fig. 11.** Documentation and Knowledge Transfer Tools

## 6.10 Retirement or Replacement Tools

Tools for the retirement or replacement of LLMs are vital for navigating their evolution amid technological advancements. Model versioning tools like Data Version Control[26] (DVC) or MLflow[27] systematically monitor changes to both data and models, creating a comprehensive history of iterations and enhancements. This documentation aids practitioners in decision-making processes related to retiring or replacing models. Continuous Integration (CI[28])/Continuous Deployment (CD[29]) pipelines, represented by tools like Jenkins or Travis CI, automate the deployment process, ensuring a seamless transition when retiring or replacing models. These tools facilitate a methodical and controlled shift, integrating new models smoothly while preserving the integrity and consistency of the overall system. Retirement or replacement tools prove indispensable for adapting to emerging requirements, surmounting challenges, and ensuring language models stay contemporary and relevant in the dynamic landscape of NLP. Through the utilization of these tools, practitioners can adeptly manage the lifecycle of LLMs, making informed decisions about when to retire or replace models to meet evolving needs and uphold optimal performance. Fig. 12 shows the block diagram related to retirement or replacement.

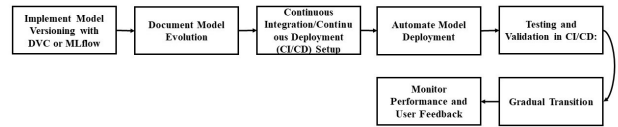

**Fig. 12.** Retirement or Replacement

## 7 COMPARISON OF LLMS TOOLS

A thorough investigation into tools for LLMs encompasses diverse aspects that are crucial for their development, deployment, and maintenance. Diverse facets critical to their development, deployment, and maintenance. In the domain of data collection and preprocessing, the versatility of Python libraries like Pandas and NumPy shines, adept at handling structured data, while NLTK proves invaluable for advanced text processing functionalities. For model development and training, TensorFlow and PyTorch emerge as leading deep learning frameworks, complemented by the user-friendly Keras and the pre-trained models from Hugging Face Transformers. Hyperparameter tuning, a pivotal step, is automated through tools like Optuna and Hyperopt, streamlining the search for optimal configurations. Evaluation and validation, essential for assessing model performance, find support in Scikit-Learn and NLTK, offering a range of metrics and tools for rigorous model assessment. Deployment tools, including TensorFlow Serving, TorchServe, and ONNX, facilitate the seamless transition of models into production environments, ensuring efficiency and reliability. Monitoring and maintenance are addressed by tools such as TensorBoard, Prometheus, and Grafana, ensuring ongoing performance and adaptability of language models. Ethical considerations and bias mitigation are actively managed through tools like AI Fairness 360 (AIF360) and Fairness Indicators, aiding in the detection and rectification of biases. User feedback and iteration tools, encompassing user analytics platforms and application-specific mechanisms, contribute significantly to refining models based on real-world experiences. Documentation and knowledge transfer leverage Jupyter Notebooks and Swagger/OpenAPI, providing comprehensive insights into model architectures and procedures. Table 1 shows the comparison of LLMs toolstack.

Table 1. Comparison of LLM Tools

| Aspect | Tools |
| --- | --- |
| Data Colleecation/Preprocceassing | Pandas, NumPy, NLTK |
| Model Development/Training | TensorFlow, PyTorch, Keras, Hugging FaceTransformers |
| Hyperparameter Tuning | Optuna, Hyperopt |

---

[23] https://jupyter.org/
[24] https://swagger.io/
[25] https://www.openapis.org/
[26] https://dvc.org/
[27] https://mlflow.org/
[28] https://about.gitlab.com/solutions/continuous-integration/
[29] https://docs.github.com/en/actions/deployment/about-deployments/about-continuous-deployment

| | |
|---|---|
| Evaluation/Validation | Scikit-Learn, NLTK |
| Deployment Tools | TensorFlow Serving, TorchServe, ONNX |
| Monitoring/Maintenance | TensorBoard, Promethes, Grafana |
| Ethical Considerations/Bias Mitigation | AI Fairness 360 (AIF360), Fairness Indicators |
| User Feedback/Iteration Tools | User analytics platforms, Application-specific mechanisms |
| Documentation/Knowledge Transfer | Jupyter Notebooks, Swagger/OpenAPI |

## 8 DISCUSSION

Although our study aims to provide a comprehensive overview of the depth and expanse of potential applications of LLMs and the tools associated with their development and deployment, it is vital to acknowledge certain drawbacks and limitations associated with it. Primary drawback pertains to the scope of our study. Despite our best efforts to entail a wide array of applications and tools, it is possible that some relevant research or tools may have been overlooked. The vast and swiftly evolving nature of the field of LLMs presents challenges in ensuring exhaustive coverage. Another potential limitation lies in the bias that may exist in our literature review. Despite our attempts at conducting a thorough review, the selection and sources may be influenced by inherent biases, which could impact the representation of certain domains or perspectives associated with LLMs. Moreover, the study illustrated in our research might not delve deeply enough into the incorporation of LLMs into fitness guidance. While it does offer a hypothetical scenario, an in-depth examination of real world application and user experience could furnish deeper insights into the challenges and opportunities associated with LLM integration in this domain. Ethical considerations are briefly addressed in our study but not fully discussed. The complex ethical implications linked to LLM development and deployment require more in-depth exploration, in addition to strategies for promotion of responsible and ethical practices. Lastly, the environmental repercussions of training and employment of LLMs, such as GPT, is a significant concern that is not extensively addressed in our study. Future researches should focus on exploring strategies for mitigating the carbon footprint associated with LLMs and encouraging environmentally sustainable practices. By acknowledging these drawbacks and limitations, we can provide a more balanced and nuanced understanding of the opportunities and challenges associated with LLMs and contribute to the advancement of responsible research and development.

## 9 CONCLUSION

The widespread deployment of LLMs across diverse realms has ushered in a new era of technological advancement and innovation. They have the potential to enhance fitness and well-being as well as transform urban planning, climate modeling, and disaster response. The capabilities of LLMs hold a promise to reshape industries and improve standard of living and quality of life. Via our thorough exploration and review of literature and tools, we have analyzed the vast research and development efforts entailing LLMs that showcase their versatility, effectiveness and potential impact. The utilization of LLMs into various domains offers transformative opportunities for personalized experiences, data-driven insights, and informed decision-making. However, as we embrace the potential of LLMs, it is crucial to take into account the various challenges pertaining to ethics, bias, privacy, and accessibility. Ethical and equitable deployment of LLMs can only be ensured with the help of responsible development practices, stakeholder engagement and interdisciplinary collaboration. Future research endeavors should attempt to address these challenges and advance LLM technology as well as explore novel applications in emerging fields. By fostering a culture of innovation, collaboration, and responsible development, we can harness the full potential of LLMs to address global challenges, drive sustainable development, and empower individuals and communities worldwide.

## DECLARATIONS

**Funding:** Not Applicable
**Conflicts of interest:** The authors declare no conflicts of interest
**Availability of data and material:** Not Applicable<.... 

**Funding:** Not Applicable
**Conflicts of interest:** The authors declare no conflicts of interest
**Availability of data and material:** Not Applicable